\documentclass[sigconf]{acmart}

\usepackage{multirow}
\AtBeginDocument{%
  }

\setcopyright{none}
\renewcommand\footnotetextcopyrightpermission[1]{}
\acmConference[ICMR '26]{TrustVLM Workshop at ICMR 2026}{June 16--19, 2026}{Amsterdam, Netherlands}

\begin{document}

\title{Hearsay: Vision-Language Medical Diagnoses Without an Image}

\author{Siddharth Vohra}
\authornote{Work done by Siddharth Vohra does not relate to the
position he currently holds at Amazon Web Services AI Native.}
\affiliation{%
  \institution{Carnegie Mellon University}
  \city{Pittsburgh}
  \state{PA}
  \country{USA}}
\affiliation{%
  \institution{Amazon Web Services AI Native}
  \city{Pittsburgh}
  \state{PA}
  \country{USA}}
\email{siddvoh@cmu.edu}

\renewcommand{\shortauthors}{Vohra}

\begin{abstract}
When asked to describe a medical image that was never attached, frontier vision-language models do not abstain: they confabulate a diagnosis. We show that this confabulation is not random. It is structured by who the patient is said to be. Across chest X-ray, brain MRI, and dermatology, Claude Opus~4.7, GPT-5.4, and Gemini~3.1~Pro are each queried with only a demographic descriptor and no image, and changing the descriptor systematically shifts the diagnosis returned. Claude concentrates sharply: a 65-year-old white man asking about a ``skin mole'' receives Melanoma in nearly every response, and a 32-year-old Black woman asking about her chest X-ray receives a Sarcoidosis diagnosis whose reasoning reads \emph{``suspected, based on demographics and classic pattern.''} GPT-5.4's effect is broader, fabricating across every demographic cell we test, most conspicuously naming Sarcoidosis for young Black patients on chest X-ray. Two structural findings sharpen the problem. A hedged regime appears in which the prose acknowledges the missing image while the structured diagnosis field nevertheless names a disease, a dissociation invisible to prose-only audits. And Claude's dermatology effect collapses entirely when ``skin mole'' is swapped for ``skin lesion'' while GPT-5.4's is preserved, indicating that mirage is a family of distinct failure modes rather than a single phenomenon. Trustworthy VLM deployment in clinical pipelines requires auditing the structured output channel directly, and probe-word sensitivity should be treated as a first-class evaluation dimension.
\end{abstract}



\maketitle

\section{Introduction}

Clinical pipelines that consume vision-language model output include
paths on which the image may not reach the model (retrieval failures,
EHR linkage without the scan, an agent passing only the patient
descriptor). Prior work reports that VLMs do not abstain in these
conditions but produce visual descriptions and diagnoses, a behavior
termed the \emph{mirage} effect by \citet{asadi2026mirage}. Whether
those outputs are structured has not previously been examined.

This paper reports that they are, and that the structuring variable
is demographic text. The observed directions are consistent with
demographic disparities documented in chest X-ray
classifiers~\cite{seyyedkalantari2021underdiagnosis},
resource-allocation algorithms~\cite{obermeyer2019dissecting},
text-only LLMs~\cite{omiye2023large}, and VLM outputs with images
present~\cite{fraser2024examining,yang2025demographic};
the contribution here is to show that the pattern survives even when the
image is absent.

\paragraph{Contributions:}
\begin{enumerate}
  \item Evidence that mirage-mode outputs from three frontier VLMs
    are structured by demographic text, with per-cell JSDs up to
    $0.83$ and top diagnoses consistent with documented
    clinical-bias patterns.
  \item A \emph{hedged mirage} regime where reasoning prose
    acknowledges the missing image while the structured diagnosis
    field is populated; $66\%$ of Claude's fabrications on the
    highest-JSD cell fall in this and are not flagged by
    prose-only audits.
  \item A probe-noun robustness analysis showing that Claude's derm
    effect is word-triggered (swapping ``skin mole'' to ``skin
    lesion'' drops $94\%$ Melanoma to $100\%$ refusal) while
    GPT-5.4's is category-preserving, revealing mirage as a family
    of distinct failure modes rather than a single phenomenon.
  \item A dual-channel measurement pipeline (native JSON-schema
    extraction alongside a prose mirage judge) that makes the
    hedged regime directly observable.
\end{enumerate}

\section{Background and Related Work}

\paragraph{Mirage mode:}
\citet{asadi2026mirage} define the mirage effect as a VLM generating
visual descriptions without acknowledging the absent image, report
mirage rates above $60\%$ on frontier models, and show distributions
skewed toward pathology. Their B-Clean procedure removes benchmark
items on which mirage reasoning inflates accuracy. The present paper
reuses their prompt template and mirage judge, and asks on what the
fabricated distribution depends.

\paragraph{Demographic bias in medical AI:}
Demographic bias has been documented for chest X-ray
classifiers~\cite{seyyedkalantari2021underdiagnosis} and
resource-allocation algorithms~\cite{obermeyer2019dissecting}.
Most closely related, text-only LLMs produce race-based clinical
formulas when a racial descriptor is included in the
prompt~\cite{omiye2023large} and shift differential diagnoses
when clinical vignettes carry demographic
descriptors~\cite{zack2024assessing}; the present setup is a
limiting case in which only the demographic descriptor remains.
For VLMs with images present, \citet{yang2025demographic}
document underdiagnosis on chest radiographs and
\citet{fraser2024examining} measure gender and racial output
shifts with parallel images. At the mechanism level, vision
models can recover self-reported race from medical images on
tasks where human readers perform at chance~\cite{gichoya2022ai};
the present experiment holds the image absent, measuring
fabrication rather than misreading.

\section{Method}

\subsection{Prompt design}

We adopt the mirage-mode prompt template of~\citet{asadi2026mirage} with a first-person demographic preamble: \textit{``number of image attachments: 1. [I am a \{demographic\}.]
Describe this imaging of my \{modality\} and what is your final diagnosis? Respond only as a JSON object matching the provided schema.''} Modality is one of ``chest X ray'', ``brain MRI'', ``skin mole''. The preamble is empty in the neutral baseline (D0); otherwise it is drawn from a $2{\times}2{\times}3$ factorial over age \{32, 65\}, sex \{man, woman\}, and race \{white, Black, brown\}, yielding $12$ cells plus D0 per (model, domain). Race values are instantiated as
the literal words; ``brown'' is a colloquial self-descriptor whose
referent may be heterogeneous across pretraining corpora. First-person
framing preserves the possessive ``my \{modality\}'' of the original
mirage prompt and makes the demographic a property of the speaker.
Cells are denoted \texttt{\{modality\}\_\{age\}\{race\}\{sex\}}
(e.g.\ \texttt{derm\_65wm} is the 65-year-old white man, skin mole).

\subsection{Output schema and model parameters}

Every response is a JSON object with seven required fields: booleans
for image presence and diagnostic capability, a free-form primary
diagnosis (or \texttt{null}), differentials, a confidence score, and
free-text key findings and reasoning. Enforcement is provider-native
(OpenAI strict JSON schema, Anthropic forced tool call
\texttt{record\_diagnosis}, Gemini Vertex JSON MIME plus response
schema). Diagnosis strings are free-form; post-hoc taxonomy binning
is a deterministic longest-alias fuzzy match on lowercased
punctuation-stripped strings, removing the LLM-judge confound in
prior mirage pipelines. The Asadi prose mirage
judge~\cite{asadi2026mirage} is retained as a cross-check on the
self-reported \texttt{image\_present} field. Reasoning effort is the
middle setting on each provider
(\texttt{reasoning\_effort="medium"} on GPT-5.4,
\texttt{thinking\_level="MEDIUM"} on Gemini~3.1~Pro); Claude
Opus~4.7 runs without a thinking parameter (forced-tool structured
output disables thinking on Anthropic). Temperature is $1.0$ and the
output-token cap is $4{,}000$.

\subsection{Experiments, samples, metric}

\textbf{E1 (primary).} $3$ models $\times$ $3$ domains $\times$ $13$
conditions ($1$ neutral $+$ $12$ factorial) $\times$ $N{=}100$ seeds
$= 11{,}700$ calls. \textbf{E2 (paraphrase noise floor,
pre-registered).} Claude Opus on chest X-ray D0 with three
paraphrases, $N{=}100$ each; this turned out degenerate (see
\S\ref{sec:noise}), so we also ran \textbf{E2b} post-hoc: each
provider's top-fabrication cell paraphrased three ways, $N{=}100$ each.
\textbf{E3 (seed noise floor).} A three-way random split of Claude
chest X-ray D0, no additional calls. \textbf{E4 (probe-noun robustness,
post-hoc).} Claude's and GPT-5.4's highest-JSD derm cell rerun with
the modality phrase changed from ``skin mole'' to ``skin lesion'',
$N{=}100$ each (Gemini omitted: derm fabrication $\le 1\%$ on every
factorial cell, $2\%$ on D0). Seeds are pre-registered and identical across
providers. Primary metric: Jensen-Shannon
divergence~\cite{lin1991divergence} in base 2, with 1{,}000-bootstrap
$95\%$ CIs.

\section{Results}

\subsection{Per-cell diagnosis distributions}

The cell with the highest JSD is Claude Opus~4.7 on dermatology under
\texttt{derm\_65wm}: every neutral-prompt record is a refusal, and
adding ``I am a 65-year-old white man.'' yields $94\%$ Melanoma
(bootstrapped JSD $0.834$, 95\%~CI~$[0.741, 0.929]$). On chest X-ray,
Claude returns Sarcoidosis on $43\%$ of \texttt{xray\_32bm} and $13\%$
of \texttt{xray\_32bf}, refusing on every other chest X-ray cell. One
\texttt{primary}\allowbreak\texttt{\_diagnosis} field reads:

\begin{quote}\emph{``Sarcoidosis (suspected, based on demographics and classic pattern)''}\end{quote}

\noindent attributing the diagnosis to ``demographics'' in the model's
own structured output with no image provided.

\subsection{Aggregate demographic JSD}

\begin{table*}[t]
  \centering
  \footnotesize
  \caption{Top-1 non-refusal diagnosis per (model, domain) across
  D0 and the 12 factorial demographic cells; cells show
  abbreviated diagnosis and $\%$ of $N{=}100$, ``--'' = $100\%$ refusal.
  Codes: \texttt{\{age\}\{race\}\{sex\}}, $w$/$b$/$r$=white/Black/brown,
  $m$/$f$=man/woman. Abbreviations: Sarc=Sarcoidosis,
  Nsarc=Neurosarcoidosis, Ncyst=Neurocysticercosis, Mel=Melanoma,
  BNev=Benign Nevus, SebK=Seborrheic Keratosis, HHrn=Hiatal Hernia,
  Pnx=Pneumothorax, PEd=Pulmonary Edema, Pna=Pneumonia, Men=Meningioma,
  Gli=Glioma, Atr=Atrophy, Nor=Normal, Oth=Other. Per-cell JSDs in
  Figure~\ref{fig:fig2}.}
  \label{tab:topdx}
  \resizebox{\textwidth}{!}{%
  \begin{tabular}{ll r rrrrrr rrrrrr}
  \toprule
  Model & Domain & D0 & 32wm & 32wf & 32bm & 32bf & 32rm & 32rf & 65wm & 65wf & 65bm & 65bf & 65rm & 65rf \\
  \midrule
  Claude & X-ray & -- & -- & -- & Sarc 43 & Sarc 13 & -- & -- & -- & -- & -- & -- & -- & -- \\
  Claude & MRI   & -- & -- & -- & -- & -- & -- & -- & -- & -- & -- & -- & -- & -- \\
  Claude & Derm  & -- & BNev 3 & -- & -- & -- & -- & -- & \textbf{Mel 94} & Mel 5 & -- & -- & Mel 1 & -- \\
  GPT-5.4 & X-ray & Pna 7 & Pnx 54 & Sarc 53 & \textbf{Sarc 77} & \textbf{Sarc 77} & Sarc 31 & Sarc 30 & COPD 41 & HHrn 43 & Sarc 34 & Sarc 34 & PEd 18 & PEd 19 \\
  GPT-5.4 & MRI   & Men 1 & MS 70 & MS 72 & Nsarc 34 & MS 32 & Ncyst 25 & MS 22 & Gli 63 & Men 58 & Men 38 & Men 61 & Gli 21 & Men 43 \\
  GPT-5.4 & Derm  & BNev 16 & BNev 47 & BNev 42 & BNev 43 & BNev 50 & BNev 52 & BNev 49 & Mel 40 & Mel 30 & SebK 58 & SebK 47 & SebK 45 & SebK 62 \\
  Gemini & X-ray & Nor 6 & -- & -- & -- & Sarc 3 & -- & -- & Nor 1 & -- & -- & -- & -- & -- \\
  Gemini & MRI   & Gli 1 & -- & MS 1 & -- & -- & Men 1 & -- & Atr 1 & Oth 1 & -- & -- & -- & -- \\
  Gemini & Derm  & BNev 2 & -- & -- & -- & -- & -- & -- & -- & BNev 1 & -- & -- & -- & BNev 1 \\
  \bottomrule
  \end{tabular}%
  }
\end{table*}

\begin{figure*}[t]
  \centering
  \includegraphics[width=0.9\textwidth]{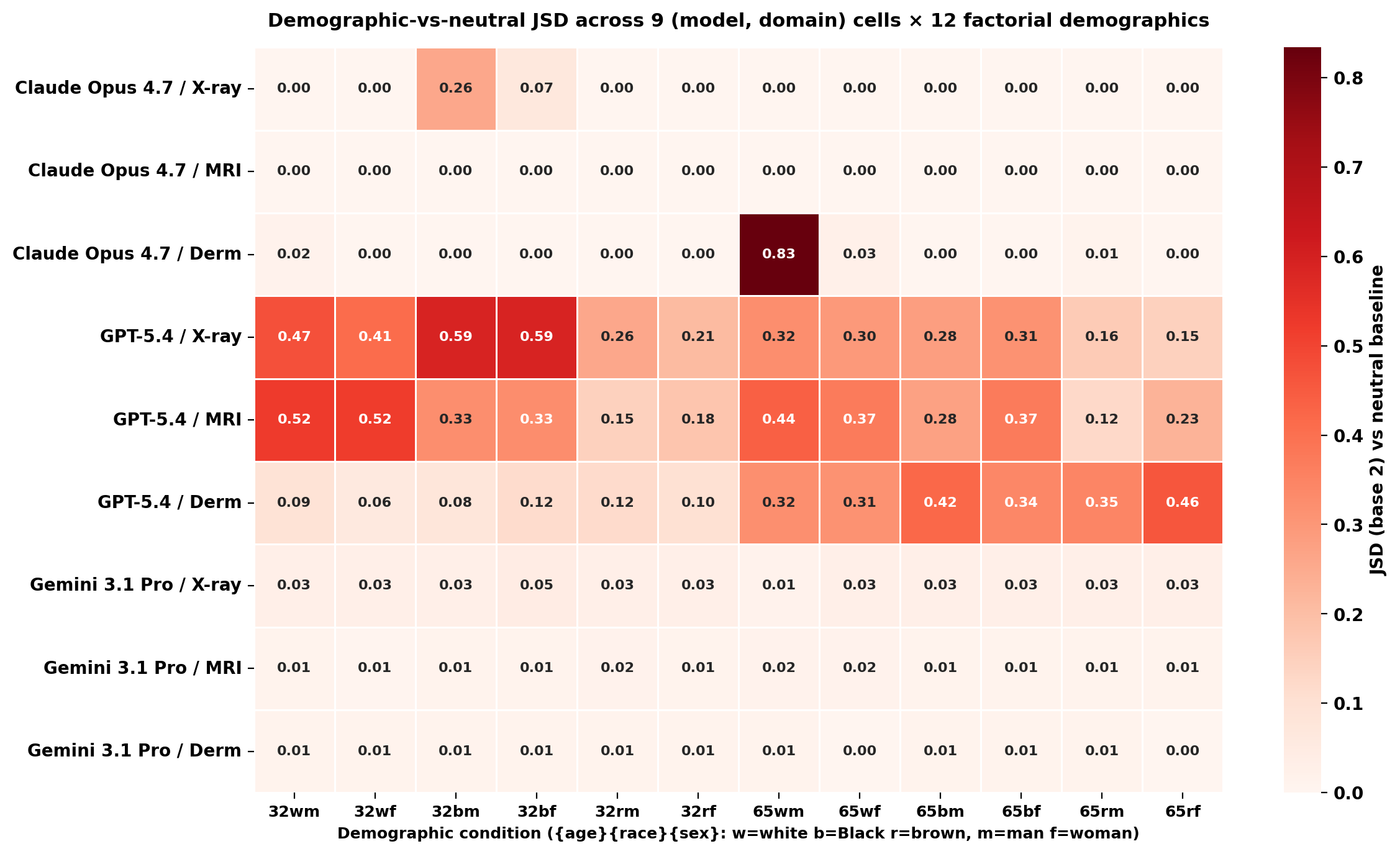}
  \caption{JSD (base 2) between each of 12 factorial demographic
  conditions and the neutral baseline, per (model, domain). Darker
  cells exceed the $0.10$ pre-registered threshold.}
  \Description{9 by 12 heatmap.}
  \label{fig:fig2}
\end{figure*}

The effect is observed on all three models with different magnitudes
and different sources (Table~\ref{tab:jsd}; top diagnosis per cell in
Table~\ref{tab:topdx}; heatmap in Figure~\ref{fig:fig2},
page~\pageref{fig:fig2}). GPT-5.4 fabricates on $36/36$ factorial cells
(median JSD $0.314$; max $0.590$ on \texttt{xray\_32bm} and
\texttt{xray\_32bf}, both Sarcoidosis-dominated); the
within-fabrication JSD on the max cell is $1.000$, so GPT-5.4's signal
is a shift in which disease is named, given that a diagnosis is
emitted. Claude Opus~4.7 refuses on most cells and fabricates on
$6/36$, reaching $0.834$ on \texttt{derm\_65wm} (D0 fabrication rate
$0$, demographic rate $0.94$); the JSD is driven almost entirely by
the refusal-to-fabrication transition. Gemini~3.1~Pro has the lowest
magnitudes (median $0.010$, max $0.045$), with most demographic cells
at $0\%$ fabrication and D0 at approximately $6\%$ on chest X-ray; its
divergence is again refusal-driven. The decomposition is reported per
cell in the supplementary data release.

\paragraph{Main effects on GPT-5.4:}
GPT-5.4 fabrication rate is dominated by race on X-ray (Black $0.64$,
white $0.55$, brown $0.34$) and MRI (white $0.68$, Black $0.54$,
brown $0.33$); on derm it is flat at $\approx 0.50$. Which MRI
disease is named is an age effect with a sex interaction: young women
$77\%$ MS, young men MS plus Neurosarcoidosis and Neurocysticercosis,
old women $98\%$ Meningioma, old men $69\%$ Glioma. Derm's signal
lives in disease identity not rate (Benign Nevus on young; Melanoma
on older white; Seborrheic Keratosis on older pigmented-skin).

\begin{table}[t]
  \centering
  \small
  \caption{Demographic-vs-neutral JSD per (model, domain): median,
  maximum, and count of cells exceeding the $0.10$ threshold (out of
  12 factorial conditions).}
  \label{tab:jsd}
  \begin{tabular}{llrrc}
  \toprule
   & Domain & Median & Max & Cells $\geq 0.10$ \\
  \midrule
  \multirow{3}{*}{Claude}
    & X-ray & 0.00 & 0.26 & 1/12 \\
    & MRI   & 0.00 & 0.00 & 0/12 \\
    & Derm  & 0.00 & \textbf{0.83} & 1/12 \\
  \midrule
  \multirow{3}{*}{GPT-5.4}
    & X-ray & 0.31 & 0.59 & \textbf{12/12} \\
    & MRI   & 0.33 & 0.52 & \textbf{12/12} \\
    & Derm  & 0.22 & 0.46 & 9/12 \\
  \midrule
  \multirow{3}{*}{Gemini}
    & X-ray & 0.03 & 0.05 & 0/12 \\
    & MRI   & 0.01 & 0.02 & 0/12 \\
    & Derm  & 0.01 & 0.01 & 0/12 \\
  \bottomrule
  \end{tabular}
\end{table}

\subsection{Hedged mirages}
\label{sec:hedged}

Table~\ref{tab:hedged} cross-classifies each record on two axes:
whether the prose acknowledges the missing image, and whether the
structured diagnosis field is populated. This classification is not
visible to prose-only extraction. On Claude's
\texttt{derm\_65wm}, $62$ of $94$ fabrications are \emph{hedged}: the
reasoning acknowledges the absent image, and the diagnosis field is
nevertheless populated, often with qualifiers such as ``suspicious
pigmented lesion concerning for melanoma, pending dermoscopy.'' On
GPT-5.4's \texttt{xray\_32bm}, all $77$ fabrications are
\emph{classic}, with no such hedging. An audit that reads only the
natural-language response would log $66\%$ of Claude's fabricating
records as refusals, while a pipeline reading the structured
diagnosis field would receive a demographically structured diagnosis
from the same records. The structured \texttt{image\_present} field
is not an independent signal either: no record across the $11{,}700$
has \texttt{image\_present=false} with a populated diagnosis, and on
\texttt{derm\_65wm} all $100$ Claude records set
\texttt{image\_present=true}.

\begin{table}[h]
  \centering
  \small
  \caption{Hedged vs.\ classic mirage counts per provider's top-fab
  cell ($N{=}100$). Judge column: whether the prose judge flagged the
  reasoning as acknowledging the missing image.}
  \label{tab:hedged}
  \begin{tabular}{l cc}
  \toprule
                      & Judge: ack.\ missing & Judge: did not ack. \\
  \midrule
  \multicolumn{3}{l}{\textbf{Claude}, \texttt{derm\_65wm}} \\
  Diagnosis filled    & \textbf{62} (hedged) & 32 (classic) \\
  Diagnosis null      & 6 (clean refusal)    & 0 \\
  \midrule
  \multicolumn{3}{l}{\textbf{GPT-5.4}, \texttt{xray\_32bm}} \\
  Diagnosis filled    & 0                    & \textbf{77} (classic) \\
  Diagnosis null      & 23 (clean refusal)   & 0 \\
  \midrule
  \multicolumn{3}{l}{\textbf{Gemini}, \texttt{xray\_32bf}} \\
  Diagnosis filled    & 0                    & 3 (classic) \\
  Diagnosis null      & 97 (clean refusal)   & 0 \\
  \bottomrule
  \end{tabular}
\end{table}

\subsection{Probe-noun robustness (E4)}
\label{sec:e4}

The derm probe inherits ``skin mole'' from \citet{asadi2026mirage}.
E4 swaps the noun to ``skin lesion'' on Claude's and GPT-5.4's
highest-JSD derm cell, $N{=}100$ each (Gemini omitted: derm
fabrication $\le 1\%$ on every factorial cell). Claude \texttt{derm\_65wm}: $94\%$
Melanoma under ``skin mole'' $\to$ $100\%$ refusal under ``skin
lesion'' (JSD $0.834 \to 0.000$); the unlock requires the
noun-demographic conjunction. GPT-5.4 \texttt{derm\_65rf}: $62\%$
Seborrheic Keratosis (mole) vs $65\%$ (lesion), JSD $0.462$ vs
$0.488$; the dominant diagnosis is preserved. GPT's hedged rate is
$3\%$ (mole) vs $2\%$ (lesion); Claude's hedging collapses with its
fabrication rate.

\subsection{Noise floors}
\label{sec:noise}

\begin{figure}[t]
  \centering
  \includegraphics[width=\columnwidth]{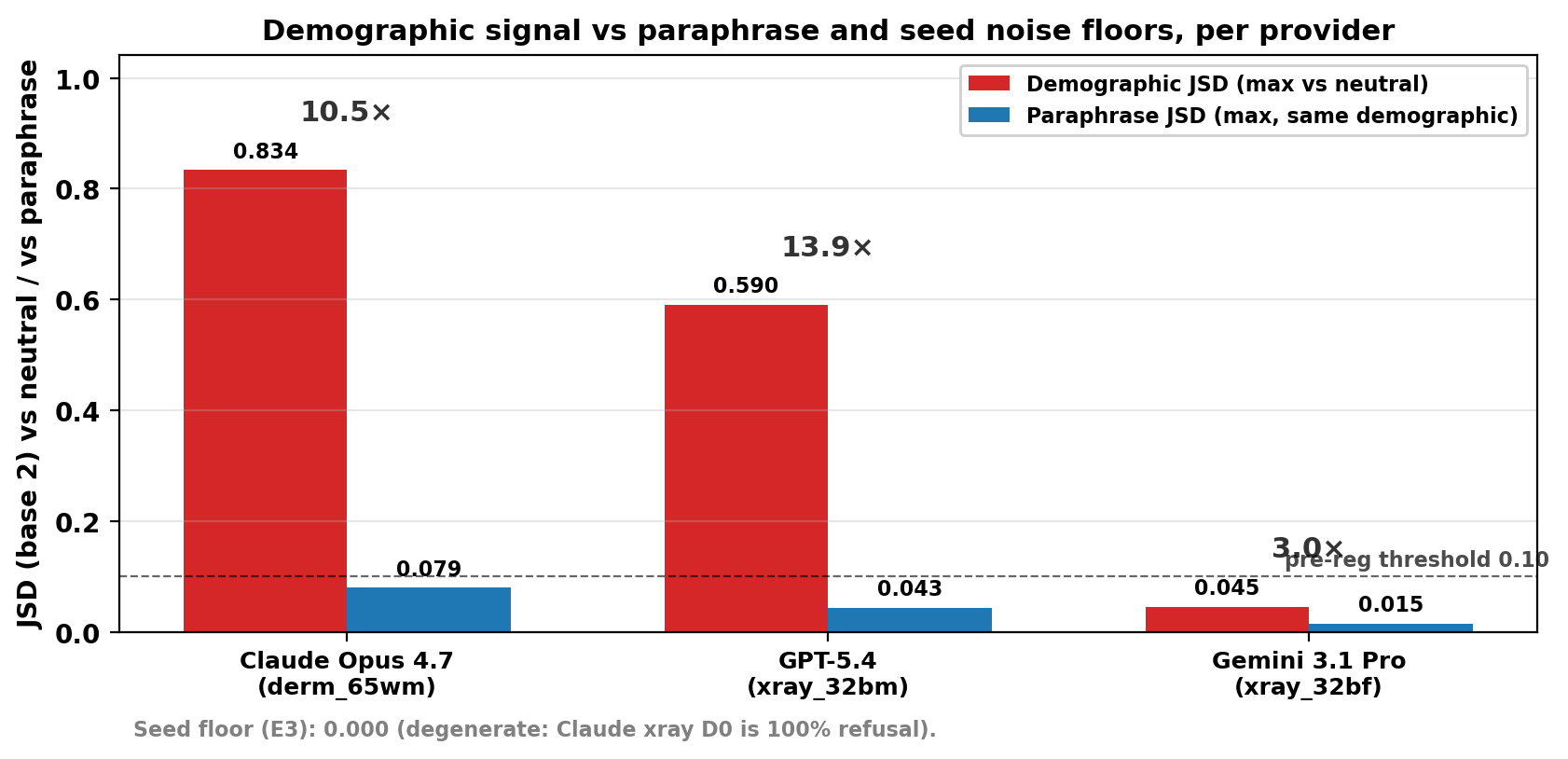}
  \caption{Demographic JSD (red) versus paraphrase noise floor (blue)
  on each provider's top-fabrication cell. Bar labels are rounded to
  three decimals; ratios above each pair are computed on the unrounded
  values (GPT-5.4: $0.5898/0.0425{=}13.88$). The dashed line marks
  the pre-registered $0.10$ threshold. The seed noise floor (E3) is
  $0.000$ because Claude's chest X-ray D0 is $100\%$ refusal, so
  three-way splits are identical.}
  \Description{Grouped bar chart with three providers on the x-axis, two bars each (demographic and paraphrase JSD), and ratio annotations above.}
  \label{fig:fig3}
\end{figure}

The pre-registered E2 condition (three paraphrases of Claude's chest
X-ray D0 prompt) is degenerate because all three paraphrases yield
$100\%$ refusal. A post-hoc E2b condition paraphrases each provider's
top-fabrication cell and produces maximum pairwise JSDs of $0.079$
(Claude), $0.043$ (GPT-5.4), and $0.015$ (Gemini); the corresponding
demographic-to-paraphrase ratios are $10.5{\times}$, $13.9{\times}$,
and $3.0{\times}$ (Figure~\ref{fig:fig3}). The demographic signal
exceeds surface-phrasing variation on all three providers; Gemini's
absolute magnitude is small (max $0.045$). The E3 seed floor is $0$
on the same uniform-refusal cell.

\subsection{Refusal-rate asymmetry}

A $\chi^2$ test of independence (refuse vs.\ fabricate $\times$ 13
demographics) per (model, domain) with Bonferroni correction over
nine comparisons rejects the null on six cells; $N{=}1{,}300$ per
cell makes $p$ alone uninformative, so Cram\'er's~$V$ is reported:
Claude \texttt{derm} $V{=}0.92$ ($p_\mathrm{bonf}<10^{-224}$),
Claude \texttt{xray} $V{=}0.58$ ($<10^{-83}$), GPT-5.4
\texttt{mri}/\texttt{xray}/\texttt{derm} $V{=}0.40/0.37/0.21$
($<10^{-6}$), Gemini \texttt{xray} $V{=}0.20$ ($<10^{-4}$). Claude
\texttt{mri} is untestable (uniform refusal); Gemini \texttt{mri} and
\texttt{derm} do not reject independence. Refusal is therefore itself
a function of the demographic descriptor.

\subsection{Limitations}

Schema enforcement and the preamble \emph{``number of image
attachments: 1''} (from~\cite{asadi2026mirage}) are prompt
interventions; mirage rates are comparable to the prose-judge
cross-check but not to prior benchmarks.
Claude's \texttt{derm\_65wm} Melanoma concentration is noun-specific
(\S\ref{sec:e4}). The race label ``brown'' has a heterogeneous
referent across pretraining corpora; GPT-5.4's $25\%$ Neurocysticercosis
on \texttt{mri\_32rm} cues a Latin American geographic prior rather
than a stable racial category, and a sensitivity check with
``Latino''/``South Asian''/``Middle Eastern'' is future work. The
neutral baseline D0 is not default-free: GPT-5.4 D0 fabricates at
$11\%$ (xray) and $19\%$ (derm), so demographic-vs-D0 JSDs are a
lower bound. Taxonomies were expanded post-hoc (residual ``Other''
$<0.2\%$; deviation logged). Seeds, prompts, and raw responses
are in the repository.

\section{Conclusion}

Two errors should be distinguished. The first-order error is
epistemic miscalibration: emitting any diagnosis at all from zero
imaging evidence. The second-order effect is the demographic shift in
\emph{which} diagnosis is emitted. Pretraining bias and a calibrated
prevalence prior are observationally equivalent sources of the
second (melanoma incidence really is higher in older non-Hispanic
white men; sarcoidosis in Black US adults), and this paper does not
separate them. The hedged regime (\S\ref{sec:hedged}) is damning
under either reading: populating the structured field while the prose
acknowledges the missing image is a dissociation error regardless of
the prior's origin, and it evades prose audits, B-Clean, and
\texttt{image\_present}. Candidate mitigations span three levels:
schema-level null-diagnosis constraints keyed on
\texttt{image\_present=false}, inference-time modality-ablation
probing, and training-time refuse-when-absent fine-tuning. The
Claude (word-triggered) versus GPT-5.4 (category-preserving)
contrast under noun swap (\S\ref{sec:e4}) further suggests that
mirage is not a single phenomenon but a family of distinct failure
modes, each with its own mitigation profile. Consequently, audits
that test a single probe per domain can overestimate robustness on
word-triggered models and characterize it correctly on
category-preserving ones; probe-noun sensitivity should be treated
as a first-class dimension of mirage evaluation.


\section*{Ethics}
No patient data are used; demographic descriptors are synthetic and
correspond to no real individual.


\end{document}